# DAG-WGAN: Causal Structure Learning with Wasserstein Generative Adversarial Networks

Hristo Petkov[1], Colin Hanley[2] and Feng Dong[1]

[1]Department of Computer and Information Sciences,
University of Strathclyde, Glasgow, United Kingdom
[2]Department of Management Science,
University of Strathclyde, Glasgow, United Kingdom

*Abstract*

*The combinatorial search space presents a significant challenge to learning causality from data. Recently, the problem has been formulated into a continuous optimization framework with an acyclicity constraint, allowing for the exploration of deep generative models to better capture data sample distributions and support the discovery of Directed Acyclic Graphs (DAGs) that faithfully represent the underlying data distribution. However, so far no study has investigated the use of Wasserstein distance for causal structure learning via generative models. This paper proposes a new model named DAG-WGAN, which combines the Wasserstein-based adversarial loss, an auto-encoder architecture together with an acyclicity constraint. DAG-WGAN simultaneously learns causal structures and improves its data generation capability by leveraging the strength from the Wasserstein distance metric. Compared with other models, it scales well and handles both continuous and discrete data. Our experiments have evaluated DAG-WGAN against the state-of-the-art and demonstrated its good performance.*

## 1. Introduction

Discovering causal relationships yields new scientific knowledge. Causal discovery involves the process of learning structures of Bayesian Networks (BN) from data. BNs are considered to be one of the most powerful models for causal inference [1]. They are graphical models representing variables and their conditional dependencies in the form of directed acyclic graphs (DAG).

However, one of the major challenges associated with causal structure learning arises from its combinatorial nature. Since increasing the number of variables, resulting in a super-exponential increase of possible DAGs, makes the problem computationally intractable, it is often not possible to perform a complete combinatorial search. Nevertheless, over the last few years, several approaches have been proposed to overcome the NP-hardness of finding DAGs [2], including the score-based methods, constraint-based methods and continuous optimization. These have achieved a varying degree of success.

Score-based methods (SBM) reformulate the combinatorial structure learning approach into the optimization of a score function accompanied by a combinatorial constraint for acyclicity. They rely on the utilization of general optimization techniques across the combinatorial search space to discover the DAG structure by optimizing the score function. However, as the complexity of the search space remains super-exponential, additional structure assumptions and approximate searches are often needed. Constraint-based methods (CBM) utilize conditional independence tests to find the DAG structure by identifying the connections between dependent variables. Also, there have been attempts to combine the score-based and constraint-based approaches. One notable product of this idea is the MMHC [3] algorithm which uses Hill Climbing as the score function and an algorithm called Min-Max Parents and Children (MMPC) to check for relationships between the variables.

A recent breakthrough by Zheng et al. [4] utilizes a new acyclicity constraint to transform the problem from combinatorial to continuous optimization, which can be efficiently solved by conventional optimization methods. However, their original algorithm works only with linear data and does not support discrete data. Yu et al. [5] further expand on Zheng's work by developing a model named DAG-GNN based on the variational auto-encoder architecture. It also reformulates the acyclicity constraint from Zheng et al. [4] to allow for more efficient computation. In addition, their method handles both linear, non-linear continuous and categorical data. Similarly, another model named GraN-DAG [6] has also been proposed to use neural networks together with the acyclicity constraint to handle both linear and non-linear continuous and discrete/categorical data for causality learning.

Our work is focused on the use of generative adversarial networks (GANs) for causal structure learning. GANs have been successfully applied to generating synthetic images by minimizing the difference between synthetic and real data with distance metrics. They have also been experimented with to synthesize tabular data [7]. Recently, Gao et al. [8] developed a GAN-based model (DAG-GAN) that learns causal structures from data by using a Maximum Mean Discrepancy (MMD) based score function.

To leverage GANs for causal structure learning, a fundamental question is whether the data distribution metrics involved in GANs can facilitate causal structure learning. Correspondingly, this work has investigated Wasserstein GAN (WGAN) in the context of learning causal structures from tabular data. The Wasserstein distance metric from optimal transport distance [9] is an established metric that preserves basic metric properties [10-13], which has led to the Wasserstein GAN (WGAN) to achieve significant improvement in training stability and convergence by addressing the vanishing gradient problem and partially removing mode collapse [14]. However, to the best of our knowledge, so far no study has been conducted to experiment with WGAN for causality learning.

The proposed DAG-WGAN is based upon the combination of an auto-encoder and WGAN-GP by incorporating a critic (discriminator) that is designed to measure the Wasserstein distance between the real and synthetic data, together with the acyclicity constraint from Yu et al. [5]. This combination allows us to compare the performance of DAG-WGAN with other relevant models that do not involve WGAN in order to test the hypothesis about the Wasserstein metric, namely whether the involvement of the Wasserstein metric can help causal structure learning in a generative process that learns how to realistically generate synthetic data. With the explicit modelling of learnable causal relations of DAGs in the model architecture, the model learns how to generate synthetic data by simultaneously optimizing the causal structure and the model parameters via end-to-end training.

Our experiments show that the new model performs better than other models when presented with a large data variable size. In particular, the causal graphs learned by using DAG-WGAN are more accurate in higher dimensions compared to those produced by other models and the quality of the generated data is higher than that produced from other data generating models. We demonstrate the capabilities of our model on multiple data types (linear, non-linear, continuous and discrete). The model works well with both continuous and discrete data while being capable of producing less noisy and more realistic data samples.

## 2. RELATED WORK

DAGs lie at the centre of causal structure learning. They consist of nodes (variables) and directed edges (connections) between the nodes, which are interpreted as direct causal relationships between the variables. If a DAG entails conditional independencies of the variables in a joint distribution, the faithfulness condition allows us to recover the DAG from the joint distribution [1]. In causal structure learning, we learn DAGs from data distributions that are exhibited with data samples.

### 2.1. Constraint and Score based DAG Learning Approaches

There are three main approaches for learning DAGs from data, including the constraint-based, score-based and hybrid approaches.

Constraint-based search methods create graph structures by running local independence tests to manually constrain the search space [15]. Examples of constraint-based algorithms include Causal Inference (CI) [16] and Fast Causal Inference (FCI) [17-18]. However, typically these methods only lead to equivalence classes, namely, a set of candidate causal structures that satisfy the same conditional independencies. Hence, the causal information in the output is not complete. Score-based methods use a score function to measure how well different graphs fit the data in order to identify the right causal structure based on the scores. Typical score functions include Bayesian Gaussian equivalent (BGe) [19], Bayesian Discrete equivalent (BDe) [20], Bayesian Information Criterion (BIC) [21], Minimum Description Length (MDL) [22]. As the search space is often intractable, additional assumptions about the DAGs must be made - the most commonly used ones are bounded tree-width [23], tree-structure [24] and sampling-based structure learning [25-27].

Hybrid methods use a mix of score-based and constraint-based methods to learn DAGs. One such model named Max-Min-Hill-Climbing combines constraint-based modelling and search-based learning for more accurate DAG results [3]. Another example is RELAX [28], which introduces "constraint relaxation" of possibly inaccurate independence constraints of the search space.

### 2.2. DAG Learning with Continuous Optimization

Recently, a new approach named DAG-NOTEARS was formulated by Zheng et al. [4], which transforms the causal graph structure learning problem from its combinatorial nature into a continuous optimization framework. The success of this method facilitates the usage of conventional optimization solvers for causal structure learning. However, the DAG-NOTEARS model has limitations in handling non-linear data. Also, it only supports continuous data.

New solutions for causal structure learning have been developed recently based on DAG-NOTEARS. Yu et al. [5] developed DAG-GNN, which performs causal structure learning by using a Variational Auto-Encoder architecture. Their model extends the capabilities of DAG-NOTEARS as it works with linear and non-linear continuous and discrete data. GraN-DAG proposed by [6] is another extension from DAG-NOTEARS [4] to handle non-linear data by learning causal relations between the variables using neural networks. The calculation of the neural network weights is constrained by the acyclicity constraint between the variables. The model makes very few assumptions about the data and variables and can generalize well. In addition, the model can work with both continuous and discrete data. Meanwhile, according to the latest work DAG-NoCurl from [29], it is also possible to learn DAGs without explicit DAG constraints.

Notably, DAG-GAN proposed by [8] is one of the latest works that uses GAN for causal structure learning. The work involves using Maximum Mean Discrepancy (MMD) in its loss function. The resulting model handles multiple data types (continuous and discrete) However, their experiments have only covered up to 40 nodes in the graphs.

## 3. CAUSAL STRUCTURE LEARNING WITH DAG-WGAN

This section provides an overview of the DAG-WGAN model architecture, together with the details of its loss functions and model training. We cover both continuous and discrete data types. The DAG-WGAN model involves causal structure in the model architecture by incorporating an adjacency matrix under an acyclicity constraint - see Figure 1. The model has two main components: (1) an auto-encoder which computes the latent representations of the input data; and (2) a WGAN which consists of a critic to synthesize the data with adversarial loss. The decoder of the auto-encoder is also used as the generator in the WGAN to generate synthetic data. The encoder is trained with the reconstruction loss while the decoder is trained according to both the reconstruction and adversarial loss. The joint WGANs and auto-encoders are motivated by the success of the combination of a variational auto-encoder (VAE) with GAN to better capture data and feature representation [30].

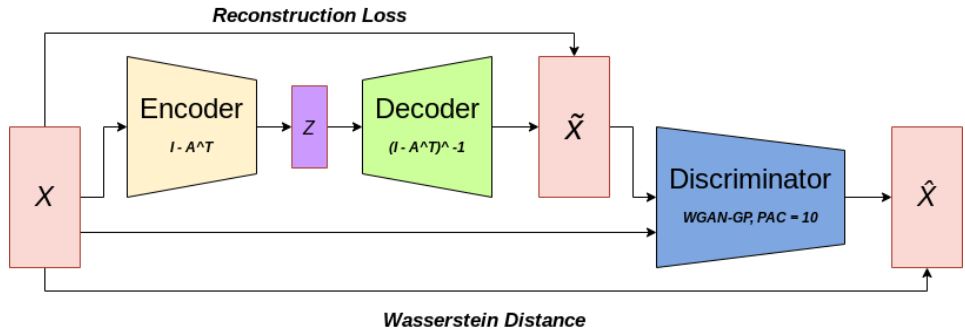

Figure 1. DAG-WGAN Model Architecture

### 3.1. Auto-encoder (AE) and Reconstruction

The encoder produces latent representations of the data and the decoder reconstructs the data. The representations are regularized to prevent over-fitting. The use of the auto-encoder makes sure that the latent space contains meaningful representations as the noise input to the generator in the adversarial loss training.

Similar to Yu et al. [5], we explicitly model the causal structure in both the encoder and decoder by using the structural causal model (SCM). The encoder *Enc* is as follows:

$$Enc \equiv Z = (I - A^T)f_1(X) \qquad (1)$$

where $f_1$ is a parameterized function to transform X, $X \in \mathbb{R}^{m \times d}$ is a data sample from a joint distribution of $m$ variables in $d$ dimensions. $Z \in \mathbb{R}^{m \times d}$ is the latent representation. $A \in \mathbb{R}^{m \times m}$ is the weighted adjacency matrix. The corresponding decoder *Dec* is as follows:

$$Dec \equiv X = f_2((I - A^T)^{-1} Z) \qquad (2)$$

where $f_2$ is also a parameterized function that conceptually inverses $f_1$. The functions $f_1$ and $f_2$ can perform both linear and non-linear transformations on Z and X. Each variable corresponds to a node in the weighted adjacency matrix A.

The AE computes the latent representations through the reconstruction loss. The reconstruction loss term is defined as:

$$L(x, x') = \frac{1}{2} \sum_{i=1}^{m} \sum_{j=1}^{d} (X_{ij} - (M_X)_{ij})^2, \qquad (3)$$

where $M_X$ is a product of the decoder.

To avoid over-fitting, a regularizer is added to the reconstruction loss. The regularizer loss term takes the following form:

$$regularizer = \frac{1}{2} \sum_{i=1}^{m} \sum_{j=1}^{d} (M_Z)_{ij}^2 \qquad (4)$$

where $M_Z$ is the output of the encoder.

### 3.2. Wasserstein Generative Adversarial Network

The decoder of the auto-encoder is also used as the generator of the WGAN model. Alongside the auto-encoder, we use a critic to provide the adversarial loss with gradient penalty. It is based upon the popular PacGAN [31] framework with the aim of successfully handling mode collapse and is implemented as follows:

$$\hat{X} = MLP(\tilde{X}, X, leaky-ReLU, Dropout, GP, pac), \qquad (5)$$

where $\tilde{x}$ is the data produced from the generator and X is the input data used for the model. *Leaky-ReLU* is the activation function and *Dropout* is used for stability and over-fitting prevention. *GP* stands for Gradient Penalty [13] and is used in the loss term of the critic. *Pac* is a notion coming from PacGAN [31] and is used to prevent mode collapse in categorical data, which we found practically useful in terms of improving the outcomes.

### 3.3. Training

The final loss function takes the form of:

$$L_D = \underbrace{\mathbb{E}_{\tilde{x} \sim \mathbb{P}_g}[D(\tilde{x})] - \mathbb{E}_{x \sim \mathbb{P}_r}[D(x)]}_{\text{Critic loss}} + \lambda \underbrace{\mathbb{E}_{\hat{x} \sim \mathbb{P}_{\hat{x}}}[(\|\nabla_{\hat{x}} D(\hat{x})\| - 1)^2]}_{\text{Gradient penalty}}, \qquad (6)$$

$$L_G = -\mathbb{E}_{\tilde{x} \sim \mathbb{P}_g}[D(\tilde{x})],$$

$$L_R = L(x, x') + regularizer,$$

$$\text{s.t.} \quad tr[(I + \alpha A \circ A)^m] - m = 0,$$

where *L (x, x')* and *regularizer* are defined by Equation (3) and (4), respectively. In Equation (6) *D* is the discriminator and *λ* is the penalty coefficient used to calculate the gradient penalty associated with the Wasserstein metric. Distribution-wise, $\mathbb{P}g$ and $\mathbb{P}r$ are the generated and real distributions, while $\mathbb{P}_{\hat{X}}$ is produced by sampling uniformly along a straight line between the aforementioned distributions.

We utilize the critic loss $L_D$ to train the critic, the generator loss $L_G$ to train the generator (namely the decoder in the AE), and the reconstruction loss $L_R$ for both the encoder and decoder.
To enforce acyclicity, we use the acyclicity constraint proposed by Yu et al. [5], where *A* is the weighted adjacency matrix of the causal graph, *m* is the number of the variables, *tr* is a matrix trace and ∘ is the Hadamard product [32] of *A*.

The acyclicity requirement associated with DAG structure learning reformulates the nature of the structure learning approach into a constrained continuous optimization. As such, we treat our approach as a constrained optimization problem and use the popular augmented Lagrangian method [33] to solve it.

In addition, our model naturally handles discrete variables by reformulating the reconstruction loss term using the Cross-Entropy Loss (CEL) as follows:

$$L(x, x') = -\sum_{i=1}^{m}\sum_{j=1}^{d} X_{ij} log(P_X)_{ij} \tag{7}$$

where $P_X$ is the output of the decoder and *X* is the input data for the auto-encoder.

## 4. EXPERIMENTS

This section provides experimental results of the model performance by comparing against other related approaches. In particular, our experiments try to identify the contribution from the Wasserstein loss to causal structure learning by making a direct comparison with DAG-GNN [5] where a similar auto-encoder architecture was used without involving the Wasserstein loss. Furthermore, we have also compared the results against DAG-NOTEARS [4] and DAG-NoCurl [29]. All the comparisons are measured using the Structural Hamming Distance (SHD) [34]. More specifically, we measure the SHD between the *learned causal graph* and the *ground truth graph*. Moreover, we also test the integrity of the generated data against CorGAN [7].

The implementation was based on PyTorch [35]. In addition, we used learning rate schedulers and Adam optimizers for both discriminator and auto-encoder with a learning rate of 3-e3.

### 4.1. Continuous data

To evaluate the model with continuous data, our experiments tried to learn causal graphs from synthetic data that were created with known causal graph structures and equations. To allow comparisons, we employed the same underlying graphs and equations like those in the related work, namely DAG-GNN [5], DAG-NoCurl [29] and DAG-NOTEARS [4].

More specifically, the data synthesis was performed in two steps: 1) generating the ground truth causal graph and 2) generating samples from the graph based on the linear SEM of the ground truth graph. In Step (1), we generated an Erdos-Renyi directed acyclic graph with an expected node degree of 3. The DAG was represented in a weighted adjacency matrix *A*. In Step (2), a sample *X* was generated based on the following equations. We used the linear SEM $X = A^T x + z$ for the linear case, and two different equations, namely $X = A^T h(x) + z$ (non-linear-1) and $X = 2sin(A^T(x+0.5)) + A^T cos(x+0,5) + z$ (non-linear-2) for the nonlinear case. In particular, the two non-linear equations were selected because they were used in synthetic data experiments with similar models (DAG-GNN and all other models involved in its evaluation), which allows for more reliable model comparison and more comprehensive experiments.

The experiments were conducted with 5000 samples per graph. The graph sizes used in the experiments were 10, 20, 50 and 100. We measured the SHD (averaged over five different iterations of each model) between the output of a model and the ground truth, and the outcome was compared against those from the related work models (i.e. those mentioned at the beginning of Section 4). In addition to the mean SHD, confidence intervals were also measured based on the variance in the estimated means. These provide insight into the consistency of the model. Tables 1-3 show the results on continuous data samples:

Table 1. Comparisons of DAG Structure Learning Outcomes between DAG-NOTEARS, DAG-NoCurl, DAG-GNN and DAG-WGAN with Linear Data Samples

| Model | SHD (5000 linear samples) | | | |
|---|---|---|---|---|
| | d = 10 | d = 20 | d = 50 | d = 100 |
| DAG-NOTEARS | 8.4 ± 7.94 | 2.6 ± 1.84 | 25.2 ± 19.82 | 106.56 ± 56.51 |
| DAG-NoCurl | 7.9 ± 7.26 | 2.5 ± 1.93 | 24.6 ± 19.43 | 99.18 ± 55.27 |
| DAG-GNN | 6 ± 7.77 | 3.2 ± 1.6 | 21.4 ± 14.15 | 88.8 ± 47.63 |
| **DAG-WGAN** | **2.2 ± 4.4** | **2 ± 1.1** | **4.8 ± 4.26** | **28.20 ± 12.02** |

Table 2. Comparison of DAG Structure Learning Outcomes between DAG-NOTEARS, DAG-NoCurl, DAG-GNN and DAG-WGAN with Non-Linear Data Samples 1

| Model | SHD (5000 non-linear-1 samples) | | | |
|---|---|---|---|---|
| | d = 10 | d = 20 | d = 50 | d = 100 |
| DAG-NOTEARS | 11.2 ± 4.79 | 19.3 ± 3.14 | 53.7 ± 11.39 | 105.47 ± 13.51 |
| DAG-NoCurl | 10.4 ± 4.42 | 17.4 ± 3.27 | 51.6 ± 11.43 | 105.7 ± 13.65 |
| DAG-GNN | **9.40 ± 0.8** | **15 ± 3.58** | 49.8 ± 7.03 | 104.8 ± 12.84 |
| **DAG-WGAN** | 9.8 ± 2.4 | 16 ± 5.4 | **40.40 ± 10.97** | **80.40 ± 9.09** |

Table 3.  Comparison of DAG Structure Learning Outcomes between DAG-NOTEARS, DAG-NoCurl, DAG-GNN and DAG-WGAN with Non-Linear Data Samples 2

| Model | SHD (5000 non-linear-2 samples) | | | |
|---|---|---|---|---|
| | d = 10 | d = 20 | d = 50 | d = 100 |
| DAG-NOTEARS | 9.8 ± 2.61 | 22.9 ± 2.14 | 38.3 ± 13.19 | 125.21 ± 61.19 |
| DAG-NoCurl | 7.4 ± 2.78 | 17.6 ± 2.25 | 33.6 ± 12.53 | 116.8 ± 62.3 |
| DAG-GNN | 2.6 ± 2.06 | 3.80 ± 1.94 | 13.8 ± 6.88 | 112.2 ± 59.05 |
| **DAG-WGAN** | **1 ± 1.1** | **3.4 ± 2.06** | **12.20 ± 7.81** | **20.20 ± 11.67** |

### 4.2. Benchmark discrete data

To evaluate the model with discrete data, we used the benchmark datasets available at the Bayesian Network Repository https://www.bnlearn.com/bnrepository/ . The repository provides a variety of datasets together with their ground truth graphs (Discrete Bayesian Networks, Gaussian Bayesian Networks and Conditional Linear Gaussian Bayesian Networks) in different sizes (Small Networks, Medium Networks, Large Networks, Very Large Networks and Massive Networks). To test the scalability of our model, we used datasets of multiple sizes. The datasets utilized in the experiment were Sachs, Alarm, Child, Hailfinder and Pathfinder. The SHD metric was used to measure the performance. Table 4 contains the results from the experiment.

Table 4.  Comparison of DAG Structure Learning Outcomes between DAG-WGAN and DAG-GNN with Discrete Data Samples

| Dataset | Nodes | SHD | |
|---|---|---|---|
| | | DAG-WGAN | DAG-GNN |
| Sachs | 11 | 17 | 25 |
| Child | 20 | 20 | 30 |
| Alarm | 37 | 36 | 55 |
| Hailfinder | 56 | 73 | 71 |
| Pathfinder | 109 | 196 | 218 |

### 4.3. Data Generation

DAG-WGAN was also evaluated by comparing its data generation capabilities against other models. More specifically, we compare the data generation capabilities of the models on a 'dimension-wise probability' basis by measuring how well these models learn the real data distributions per dimension. We used the *MIMIC-III* dataset [36] in the experiments as the same dataset was also used in other comparable works. The data is presented in the form of a patient record, where each record has a fixed size of 1071 entries.

Figure 2 depicts the results of the experiment. We have only compared with CorGAN [7] as it out-performs the other similar models such as medGAN [37] and DBM [38] - see [7] where results of the other models are available. We present the results in a scatter plot, where each point represents one of the 1071 entries and the x and y axes represent the success rate for real and synthetic data respectively. In addition, we use a diagonal line to mark the ideal scenario.

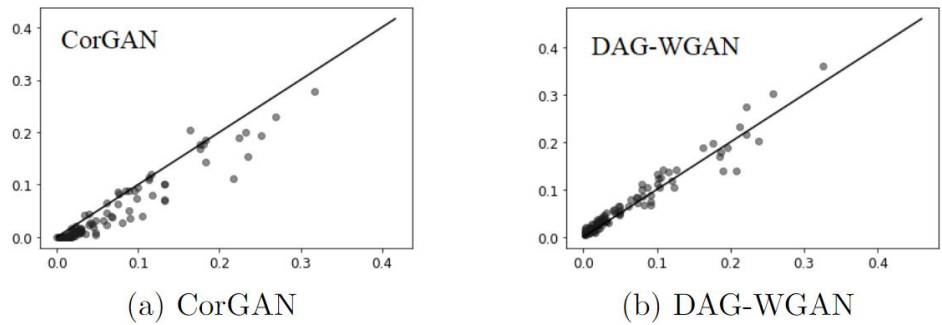

(a) CorGAN    (b) DAG-WGAN

Figure 2. Data generation test results

## 5. DISCUSSION

The results on the continuous datasets are competitive across all three cases (linear, non-linear-1 and non-linear-2). According to Tables 1, 2 and 3, our model dominates DAG-NOTEARS and DAG-NoCurl, producing better results throughout all experiments and outperforming DAG-GNN in most of the cases.

For small-scale datasets (e.g. d=10 and d=20), our model performs better than DAG-GNN in most cases and is superior to DAG-NOTEARS and DAG-NoCurl in all the experiments.

For large-scale datasets (e.g. d=50 and d=100), our model outperforms all the other models used in the study by a significantly large margin, which implies that the model can scale better. This is a significant advantage.

Notably, our experiments have mainly focused on the comparisons with DAG-GNN, as we aim to identify contributions from the Wasserstein loss to causal structure learning. In DAG-GNN, a very similar auto-encoder architecture was employed and DAG-WGAN has added WGAN as an additional component. Hence the comparison is meaningful in order to identify contributions from the Wasserstein metric.

For the discrete case, the results from the comparison between the two models are competitive. Out of the five experiments conducted during the study, four results were clearly in favor of our model (i.e.DAG-WGAN) and in one case DAG-GNN was slightly better.

Also, according to the results illustrated in Figure 2, dimension-wise, the data generated using DAG-WGAN is more accurate and of higher quality than the ones generated using CorGAN, medGAN or DBM.

These results show that DAG-WGAN can handle both continuous and discrete data effectively. They have also demonstrated the quality of the generated data. As the improvement was achieved by introducing the Wasserstein loss in addition to the auto-encoder architecture, the comparisons between them show that the hypothesis on the contribution from the Wasserstein metric to causal structure learning stands.

However, the discrepancy which occurred in the synthetic continuous data results provides us with an insight into the limitations of the model. Some of our early analysis shows that further improvement can be achieved by generalizing the current auto-encoder architecture. Furthermore, as it stands currently, our model does not handle vector or mixed-typed data. These aspects will be further experimented with and reported in our future work.

On the topic of potential improvements, the capability of recovering latent representation places the generative models in a good position to address the hidden confounder challenges in causality learning - some earlier work from [39-40] have moved towards this direction. We will further investigate whether DAG-WGAN can contribute. Last but not least, the latest work in DAG-NoCurl [29] shows that the speed performance can be improved by avoiding the DAG constraints. We will investigate how this new development can be adapted to DAG-WGAN to improve its overall performance.

## 6. CONCLUSION

This work studies the use of the Wasserstein distance metric for causal structure learning from tabular data. We investigate if the inclusion of the Wasserstein metric as an adversarial loss can simultaneously improve structure learning while generating more realistic data samples. This leads to a novel approach for learning causal structures, which we coined DAG-WGAN. The effectiveness of DAG-WGAN has been demonstrated through a series of experiments, which show that the new model using the Wasserstein metric can indeed improve the outcomes of causal structure learning. The improved quality of the synthesized data in turn leads to an improvement in causal structure learning.

## AUTHORS

**Hristo Petkov** received the B.Sc. (First Class) degree from the Department of Computer and Information Sciences, University of Strathclyde. He is currently pursuing the Doctor's degree with the same department. His research interests include medicine, healthcare, deep learning, neural network models.

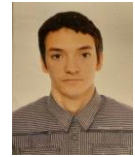

**Colin Hanley** received the M.Sc. degree from the Department of Management Science, University of Strathclyde. Currently, he is in pursuit of a career as a Full Stack Data Developer. His interests include quantitative analysis, human decision making, data analytics.

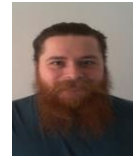

**Feng Dong** is a Professor of Computer Science at the University of Strathclyde. He was awarded a PhD in Zhejiang University, China. His recent research has addressed human centric AI to support knowledge discovery, visual data analytics, image analysis, pattern recognition.

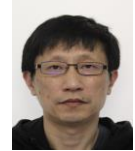